\documentclass{article}
\usepackage{akbc}
\akbcheading
\ShortHeadings{Iterative Knowledge Graph Refinement with Symbolic and NeuralMethods}{Anonymous}
\usepackage{times}
\usepackage{soul}
\usepackage{url}
\usepackage[utf8]{inputenc}
\usepackage{graphicx}
\usepackage{amsmath}
\usepackage{amsthm}
\usepackage{booktabs}
\usepackage{algorithm}
\usepackage{algpseudocode}
\algnewcommand{\LineComment}[1]{\State \(\triangleright\) #1}
\usepackage[table]{xcolor}
\usepackage[draft]{todonotes}   

\usepackage{latexsym}
\usepackage{multirow}
\usepackage{booktabs}
\usepackage{amsmath}
\usepackage{paralist}
\usepackage{xspace}
\usepackage{xcolor,colortbl}

\newcolumntype{a}{>{\columncolor{pink!15}}c}

\newcommand\comment[1]{}

\newcommand{\complex}{ComplEx\xspace}
\newcommand{\conve}{ConvE\xspace}
\newcommand{\typeex}{TypeE-X\xspace}
\newcommand{\typeecomplex}{TypeE-ComplEx\xspace}
\newcommand{\typeeconve}{TypeE-ConvE\xspace}
\newcommand{\pslkgi}{PSL-KGI\xspace}

\newcommand{\fb}{FB15K-237\xspace}
\newcommand{\yago}{YAGO3-10\xspace}
\newcommand{\wnet}{WN18RR\xspace}

\newcommand{\nell}{NELL\xspace}

\newcommand{\ourm}{{IterefinE}\xspace}

\finalcopy 
\begin{document} 
\title{IterefinE: Iterative KG Refinement Embeddings using Symbolic Knowledge}
\author{\name Siddhant Arora \email siddhantarora1806@gmail.com \\
       \name Srikanta Bedathur \email srikanta@cse.iitd.ac.in \\
       \name Maya Ramanath \email ramanath@cse.iitd.ac.in \\
       \name Deepak Sharma \email dsharma080@gmail.com\\
}


\maketitle
\begin{abstract}
Knowledge Graphs (KGs) extracted from text sources are often noisy and lead to poor performance in downstream application tasks such as KG-based question answering. While much of the recent activity is focused on addressing the sparsity of KGs by using embeddings for inferring new facts, the issue of cleaning up of noise in KGs through \emph{KG refinement} task is not as actively studied. 
Most successful techniques for KG refinement make use of inference rules and reasoning over ontologies. Barring a few exceptions, embeddings do not make use of ontological information, and their performance in KG refinement task is not well understood. In this paper, we present a KG refinement framework called \textbf{\ourm} which iteratively combines the two techniques -- one which uses ontological information and inferences rules, {\it viz.,}\pslkgi, and the KG embeddings such as \complex and \conve which do not. As a result, \ourm is able to exploit not only the ontological information to improve the quality of predictions, but also the power of KG embeddings which (implicitly) perform longer chains of reasoning. The \ourm framework, operates in a co-training mode and  results in \emph{explicit type-supervised} embeddings of the refined KG from \pslkgi which we call as \textbf{\typeex}. Our experiments over a range of KG benchmarks show that the embeddings that we produce are able to reject noisy facts from KG and at the same time infer higher quality new facts resulting in upto 9\% improvement of overall weighted F1 score. 
\end{abstract}

\section{Introduction}
Knowledge graphs (KGs) represent facts as a set of directed edges or triples $\langle$\texttt{s,r,o}$\rangle$ where \texttt{r} is the relation between entities \texttt{s} and \texttt{o}. 
	A critical issue in large-scale KGs is the presence of noise from the automatic extraction methods used to populate them. For instance, NELL \cite{carlson2010toward} is known to contain various kinds of errors including: different names for the same entity (e.g., \texttt{australia} and \texttt{austalia}), incorrect relationships --both due to wrong relation label as well as incorrect linkage altogether-- between entities (e.g., $\langle$\texttt{matt\_flynn, athleteplayssport, baseball}$\rangle$ is \textit{false} since Matt Flynn is an NFL player), incompatible entity types, and many more~\cite{pujara2013knowledge}. It has also been observed that such noise can significantly degrade the performance of KG embeddings~\cite{pujara2017sparsity}. 

The \emph{KG refinement} task aims to reduce the noise in KG by not only predicting additional links (relations) and types for entities (i.e., performing \emph{KG completion}), but also eliminating incorrect facts. Methods for noise reduction in KG include the use of association rule mining over the noisy KG to induce rules which can help in eliminating incorrect facts~\cite{ma2014learning}; reconciling diverse evidence from multiple extractors~\cite{dong2014knowledge}; the use of ontology reasoners~\cite{NakasholeTW11} and many more. A detailed survey of approaches for KG refinement is available in~\cite{paulheim2017knowledge}. On the other hand, neural and tensor-based embeddings have seen significant success in entity type and new fact predictions~\cite{nickelyago,trouillon2016complex,dettmers2018convolutional}. It is worth noting that embeddings, with a few recent exceptions~\cite{guoKale2016,minervini2017,minerviniNTP2018,fatemi2019aaai}, do not make use of rich taxonomic/ontological rules when available. Methods such as Probabilistic Soft Logic (PSL) and Markov Logic Network (MLN) have been adapted for the KG refinement problem, and can address both the completion as well as noise removal stages of the KG completion problem. They can also make use of ontological rules effectively, and specifically, the \pslkgi implementation uses rules defined on schema-level features~\cite{pujara2013knowledge}.

\subsection{Contributions} In this paper we investigate the combined use of ontologies and embeddings in the KG refinement task. Ontologies are among the best methods to eliminate noisy facts in KGs, while embeddings provide a means of \emph{implicitly} reasoning over longer chains of facts. Specifically, we use Probabilistic Soft Logic (PSL) that can incorporate inference rules and ontologies, along with state-of-the-art KG embedding methods,{\it viz.,} ConvE \cite{dettmers2018convolutional} and ComplEx \cite{trouillon2016complex}, which do not make use of any ontological rules. 

The resulting framework called \ourm is based on the observation that the mispredictions by the embeddings based methods are often due to the lack of type compatibility between the entities due to their type-agnostic nature~\cite{xie2016representation,jain2018type}. Since \pslkgi is able to predict entity types by making use of ontological information along with many candidate facts derived using its inference rules, \ourm transfer these predictions from \pslkgi to the embeddings. This results in embeddings with \emph{explicit} type supervision, which we call as \textbf{\typeecomplex} and \textbf{\typeeconve}. Further, we feed the predictions back from \typeecomplex (correspondingly, \typeeconve) over the training set to the \pslkgi, resulting in additional evidence for inference. This feedback cycle can be repeated for multiple iterations, although we have observed over various benchmark datasets that the performance stabilizes within 2 to 3 iterations. Our key findings reported in this paper are as follows: 
\begin{enumerate}[(i)] 
\item Explicit type supervision improves the weighted F1-score of embeddings by up to $9\%$ over those which do not have type supervision.
\item Explicit type supervised models also outperform the implict type supervised models~\cite{jain2018type}. The margin of improvement is large when the ontological information is sufficiently rich to begin with. 
\item Rich ontological information is a critical ingredient for the performance of \typeeconve and \typeecomplex, particularly when we consider their ability to remove the noisy triples. We observed that on datasets like \yago  and \fb, we improved F1 scores on noisy triples by 30\% to 100\%.
\end{enumerate}
We note that, although we have experimented with \conve and \complex, it easy to instantiate \ourm to work with other embeddings, which we plan to explore in our future work. 



\section{Related Work}
\label{sec:relwork}
 In this section, we describe how KG refinement is accomplished by methods based on inference rules and embeddings-based methods. There are other research directions for (partially) solving the KG refinement problem such as rule induction~\cite{ma2014learning}, classification with diverse extractors~\cite{dong2014knowledge}, crowdsourcing, etc., (see \cite{paulheim2017knowledge} for an overview). While these works have their own strengths and weaknesses, our focus in this paper is on the use of ontological rules (exemplified by \pslkgi) and embeddings (we use ComplEx, ConvE and \cite{jain2018type}). Rule induction methods are orthogonal to our work, and may augment or replace the set of rules we use. Further, evidence from diverse extractors as in the case of \cite{dong2014knowledge} can be incorporated into the \pslkgi framework in a straightforward manner (see details about confidence values of triples in the Background section).


\subsection{KG Refinement with Ontological Rules} 
\label{sec:pslkgi}
Methods based on Markov-Logic Networks or Probabilistic Soft Logic (PSL), model the KG refinement task as a constrained optimization problem that scores facts in the KG with the help of various symbolic (logical) rules. An important input to these formulations are the probabilistic sources of information such as the confidence scores obtained during extraction~\cite{pujara2013knowledge,jiang2012learning} from multiple sources. 

Of these methods, \pslkgi~\cite{pujara2013knowledge,pujara2017sparsity} is shown not only to perform better with KG noise and sparsity, but also to be quite scalable. It uses the following sources of information in addition to the noisy input KG: confidence scores of extractions, a small seed set of manually labeled correct facts and type labels and ontology information and inference rules.

\comment{
\begin{table}[h]
\centering
\resizebox{\columnwidth}{!}{\begin{tabular}{l}
\toprule
$SameEntity(A,B) \wedge Cat(A,C) \rightarrow Cat(B,C)$ \\
$SameEntity(A,B) \wedge Rel(A,Z,R) \rightarrow Rel(B,Z,R)$ \\
$SameEntity(A,B) \wedge Rel(Z,A,R) \rightarrow Rel(Z,B,R)$ \\
$Sub(C,D) \wedge Cat(A,C) \rightarrow Cat(A,D)$ \\  
$RSub(R,S) \wedge Rel(A,B,R) \rightarrow Rel(A,B,S)$ \\ 
$Mut(C,D) \wedge Cat(A,C) \rightarrow \neg Cat(A,D)$ \\
$RMut(R,S) \wedge Rel(A,B,R) \rightarrow \neg Rel(A,B,S)$ \\
$Inv(R,S) \wedge Rel(A,B,R) \rightarrow Rel(B,A,S)$ \\
$Domain(R,C) \wedge Rel(A,B,R) \rightarrow Cat(A,C)$ \\
$Range(R,C) \wedge Rel(A,B,R) \rightarrow Cat(B,C)$ \\
\bottomrule
\end{tabular}
}
\caption{The List of Inference Rules used by \pslkgi~\cite{pujara2013knowledge}.}
\label{rules}
\end{table}
}

\subsection{Refinement task with KG embeddings}
KG embedding methods define a scoring function $f$ to score the plausibility of a triple\footnote{See \cite{wang2017knowledge} for a survey of embedding methods and the many forms the scoring function $f$ can take.} and learn embeddings in such a way as to maximise the plausibility of the triples that are already present in the KG~\cite{nickel2011three,socher2013reasoning,trouillon2016complex}.

An important step in learning is the generation of negative samples since the existing triples are all labeled positive. The negative samples are typically generated by corrupting one or more components of the triple. With this dataset containing both positive and negative samples, training can be done for the refinement task with a negative log-likelihood loss function as follows \cite{trouillon2016complex}.

\begin{align}
\begin{split}
L(G) = \sum_{(s,r,o,y) \in G} &  y \log{f(s,r,o)} + \left(1-y\right) \log{(1-f(s,r,o))}
\end{split}
\end{align}
%
%
%
%
where $(s,r,o)$ is the relation triple, $f$ is the scoring function, and $y$ denotes whether the triple is given positive label or negative.
Similar to the setting for \pslkgi, embedding-based methods can also be used to predict type labels of entities (the \emph{typeOf} relation). We work with \complex ~\cite{trouillon2016complex} and \conve~\cite{dettmers2018convolutional} embeddings which have shown state of the art performance in many KG prediction tasks.

\subsection{Type and Taxonomy Enhanced Embeddings}
There are some recent efforts to incorporate type hierarchy information in KG embeddings --e.g., TKRL~\cite{xie2016representation} and TransC~\cite{lv-etal-2018-differentiating}. 
Recently, SimplE$^+$~\cite{fatemi2019aaai} includes taxonomic information --i.e., subtype and subproperty information-- and also shows that state-of-the-art embeddings like ComplEx~\cite{trouillon2016complex}, SimplE~\cite{kazemi2018}, ConvE~\cite{dettmers2018convolutional} cannot enforce subsumption. 

Taking a different approach~\cite{jain2018type} propose extending standard KG embeddings \emph{without explicit} type supervision by representing entities as a two-part vector  with one part encoding only the type information while the other one is a traditional vector embedding of the entity (and corresponding change to the relation embeddings as well). Specifically it uses the following scoring function : 
\begin{align}
\begin{split}
f (s, r, o) = \sigma(\mathbf{s_t}\cdot\mathbf{r_h}) * \mathbf{Y} (s, r, o) * \sigma(\mathbf{o_t} \cdot \mathbf{r_t}),
\end{split}
\end{align}
where $\mathbf{s_t}$ and $\mathbf{o_t}$ denote the embedding vectors for implicit type label of entities, and $\mathbf{r_{h}}$ and $\mathbf{r_{t}}$ denote the implicit type embeddings for domain and range of relation $r$. $\mathbf{Y}$ is the scoring function used by the underlying embeddings-based method -- we experiment with ComplEx and ConvE.

These embeddings enforce type compatibilities during KG link prediction task, and they showed nearly 5-8 point improvements in MRR and type F1 scores. In our work, we build on this idea further by adding another layer of explicitly supervised type vector to entity/relation embeddings. 

Note, however, that our focus in this paper is not on embeddings that enforce ontological constraints, but on improving the KG refinement by combining the strengths of KG embeddings with methods like \pslkgi and MLNs which can work with arbitrary (first-order) constraints.\\

Recently, there has been some work in modeling structural as well as uncertainty information of relations in the embedding space. \cite{sun2019uncertain} uses Probabilistic Soft Logic to come up with plausibility scores for each fact which they train to match with the uncertainty score of seen relation triplets as well as minimize the plausibility score for relation triplets. However, they do not focus on the KG refinement task and they also do not investigate how existing Knowledge Graph Embedding methods can be used in conjunction with this approach to effectively embed Uncertain graphs. There has also been some research in using rule-based reasoning and KG embeddings together in an iterative manner in \cite{wen2019iterative}. They achieve improvements in the performance of link prediction tasks for sparse entities which cannot be effectively modelled by standard embedding methods. However, at each iteration, they are adding more rules to their database, which makes their approach less scalable to us since we are continuously removing noise from Knowledge Graph, thus making the size of resultant Knowledge Graph stable. Also, the feedback in their work was rules learned from embedding with a robust pruning strategy. In contrast, we passed feedback as relation triples along with their predicted score as additional context for the \pslkgi model to generate high quality predictions. Finally, we test this feedback in Knowledge Graph refinement manner where we couple the task of removing noise as well as inferring new rules together in a coupled manner with both the tasks benefiting from each other.

\section{Background}
\label{sec:background}

We use the \pslkgi implementation generously provided by the authors\footnote{\url{https://github.com/linqs/psl-examples/tree/master/knowledge-graph-identification}}.
The inputs to \pslkgi are: 
\begin{enumerate}[(i)]
	\item the triples extracted from multiple input sources and confidence values for these triples,
	
	\item ontology information, such as sub-class (SUB) and sub-property (RSUB) information; the domain and range of relations (DOM, RNG); "same" entities (SAMEENT), entities and relations that are mutually exclusive (MUT and RMUT); and inverse relations (INV). We reproduce the list of  information used in \cite{pujara2013knowledge} in tabular form in Table \ref{tab:onto_info}.
	
	\item Inference rules -- specifically, there are 7 general constraints that were first introduced in the earlier work on Markov Logic Networks (MLN) based work~\cite{jiang2012learning}. These rules are listed in Appendix \ref{sec:appendix} in Table ~\ref{onto_rules}. 
\end{enumerate}

\begin{table}[h]
\centering
\begin{tabular}{ll}
\toprule
Ontological Information & Description \\
\midrule
Domain (DOM) & Domain Of relation  \\
Range (RNG)& Range of relation \\
Same Entity (SAMEENT) & Helps perform Entity Resolution by specifying equivalence class of entities  \\
MUT & Specifies that 2 entities are mutually exclusive in their type labels \\
Subclass (SUB) & Subsumption of labels \\
INV & Inversely related relations \\
RMUT & Mutually exclusive relations   \\
SUBPROP (RSUB) & Subsumption of relations \\

\bottomrule
\end{tabular}
\caption{Ontological Information used in PSL-KGI Implementation}
\label{tab:onto_info}
\end{table}

\begin{table}
\centering
\begin{tabular}{lc}
\toprule
Class & Ontological Rule \\
\midrule
\multirow{2}{*}{Uncertain Extractions} & $w_{CR-T} : CANDREL_{T}(E_1,E_2,R) \Rightarrow REL(E_1,E_2,R)$   \\
 & $w_{CL-T} : CANDLBL_{T}(E,L) \Rightarrow LBL(E,L)$ \\
 \cline{1-2} \\
\multirow{3}{*}{Entity Resolution} & $SAMEENT(E_1,E_2) \land LBL(E_1,L) \Rightarrow  LBL(E_2,L)$  \\
& $SAMEENT(E_1,E_2) \land REL(E_1,E,R) \Rightarrow  REL(E_2,E,R)$ \\
& $SAMEENT(E_1,E_2) \land REL(E,E_1,R) \Rightarrow  REL(E,E_2,R)$ \\
INV & $INV(R,S) \land REL(E_1,E_2,R) \Rightarrow REL(E_2,E_1,S)$ \\
\cline{1-2}\\
\multirow{2}{*}{Selectional Preference} & $DOM(R,L) \land REL(E_1,E_2,R) \Rightarrow LBL(E_1,L)$   \\
 & $RNG(R,L) \land REL(E_1,E_2,R) \Rightarrow LBL(E_2,L)$ \\
 \cline{1-2}\\
\multirow{2}{*}{Subsumption} & $SUB(L,P) \land LBL(E,L) \Rightarrow LBL(E,P)$  \\
& $RSUB(R,S) \land REL(E_1,E_2,R) \Rightarrow REL(E_1,E_2,S)$ \\ 
\cline{1-2}\\
\multirow{2}{*}{Mutual Exclusion} & $MUT(L_1,L_2) \land LBL(E,L_1) \Rightarrow \neg LBL(E,L_2)$  \\
& $RMUT(R,S) \land REL(E1,E2,R) \Rightarrow \neg REL(E1,E2,S)$ \\

\bottomrule
\end{tabular}
\caption{Ontological Inference Rules used by PSL-KGI}
\label{onto_rules}
\end{table}

Based on these \pslkgi defines a PSL program that combines the ontological rules and constraints with atoms in the KG.The ontological information and inference rules are summarized in Table \ref{tab:onto_info} and Table \ref{onto_rules} respectively. 
The solution to the PSL program essentially provides most likely interpretation of the KG, defining a probability distribution over the KG. By appropriately selecting the threshold on the probability value, it is possible to reject noisy facts. It is also important to note that \pslkgi also generates a number of candidate facts that are not originally in the KG by soft-inference over the ontology and inference rules. While the extraction confidence for a triple may be high, it is possible for \pslkgi to output a low score for that triple because of the inference rules. This enables it to determine correct type labels and in expanding the seed set iteratively.

\section{Combining PSL-KGI with KG embeddings}
\subsubsection*{Overview}
\begin{figure}[h]
    \centering
    \includegraphics[width=0.5\textwidth,bb=0 0 1020 540]{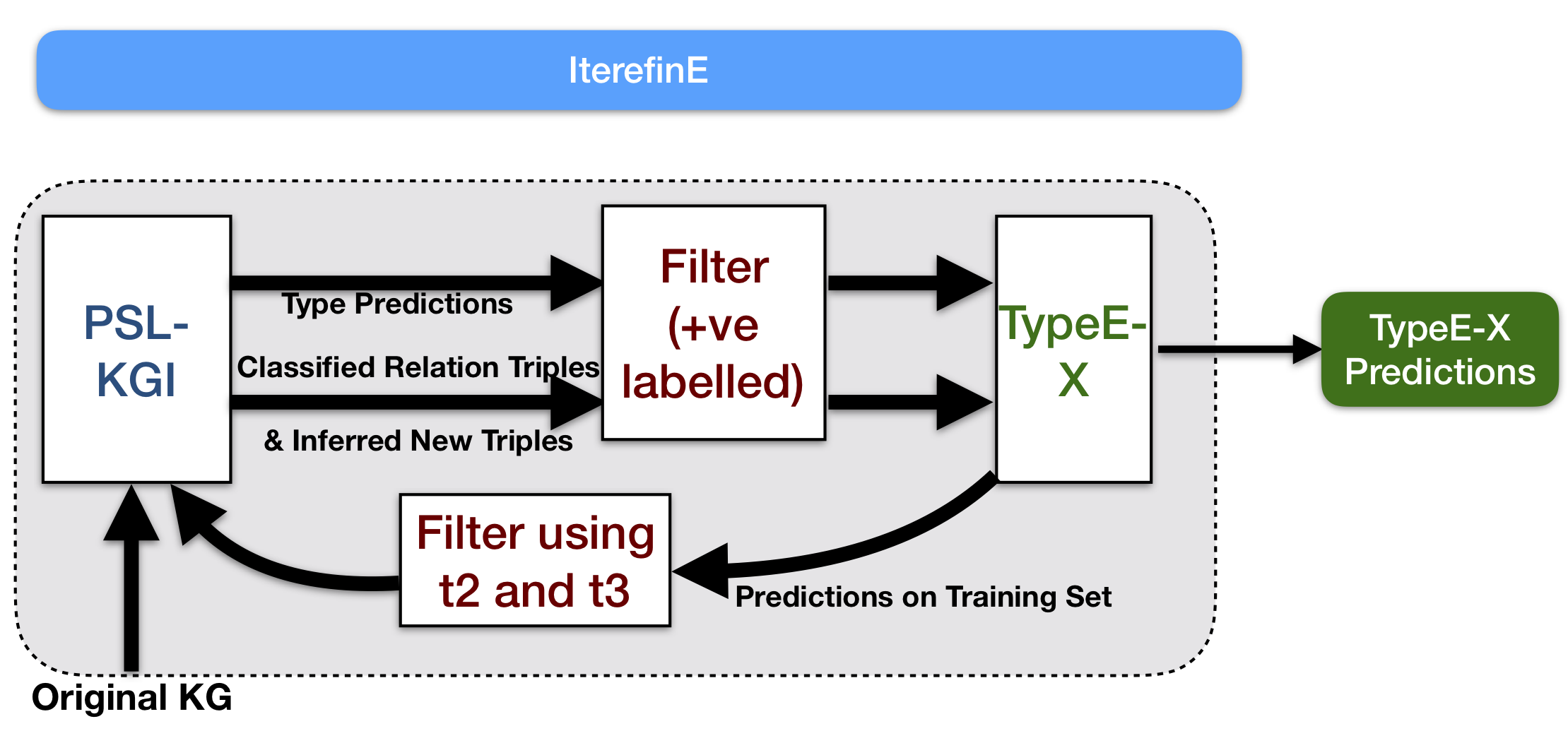}
    \caption{\ourm: Combining \pslkgi and embedding model X resulting in \typeex model.}
    \label{combined_model}
\end{figure}

We now present a simple mechanism, partially based on the concept of co-training~\cite{blum1998combining}, to combine the strengths of \pslkgi and KG embeddings. The mechanism consists of two stages, as shown in Figure~\ref{combined_model}. In the first stage, \pslkgi is used to generate high-quality type predictions, and in the second stage, an enhanced KG embeddings method, which we term as \typeex (where \textbf{X} is an embeddings method such as \complex), takes as input, the type predictions and relation triples labeled \emph{true} in the training set. At the end of the second stage, the embeddings generated are expected to be of higher quality. 
%
%
The feedback to \pslkgi is completed by passing the predictions from the KG refinement of \typeex back to \pslkgi which takes them, along with the original extraction scores, as additional context for predicting relation triples. 
Note that this process can be repeated iteratively, allowing the propagation of potentially more context at each iteration\footnote{For an algorithmic listing of \ourm, please refer to Appendix \ref{sec:psuedo_algo}}.

Our observations show that passing \emph{all} newly predicted triples by \typeex back to \pslkgi as feedback would make our approach nonscalable for multiple iterations. Therefore, we only add some of the top most positive and most negative relations so that the size of the KG remains stable without sacrificing accuracy.


In order to ensure that an optimal number of positive and negative triples are fed back to \pslkgi, we calculate \emph{separate} thresholds for each. First, the classifier threshold $t_1$ determines which triples are predicted as positive and which are negative. This threshold is determined by optimizing over a validation set. Second, we divide the set of triples, using $t_1$, into positive triples, denoted by $P_1$, and negative triples, denoted by $N_1$.  Now, we choose two new thresholds $t_2$ and $t_3$:
\begin{equation}
\begin{split}
    t_2 = t_1+\Phi_1*mean(P_1)\\
    t_3 = t_1-\Phi_2*mean(N_1)
\end{split}
\label{eqn:thresholds}
\end{equation}

where $mean(X)$ is the mean score of triples in set $X$, $\Phi_1$ and $\Phi_2$ are parameters that can be tuned. Then we add all relations with predicted probability greater than $t_2$, along with their inferred probabilities, as a form of positive feedback for our PSL-KGI model of the next iteration. Similarly, the negative feedback would consist of all relations with predicted probabilities less than $t_3$. We discuss the impact of these thresholds on the size of the KG and the prediction accuracy in Section \ref{sec:hyper-parameters}. 

\subsubsection*{Scoring function for \typeex}
To incorporate the type inferences for entities generated by \pslkgi in KG embeddings (the second stage), we modify the \emph{typed model}~\cite{jain2018type} as follows:

Instead of just using the implicit type embeddings, we concatenate them with embeddings of explicit types transferred from \pslkgi. Note that the implicit type embeddings are learned for each entity or relation, whereas the explicit type embeddings are the same for all entities with the same type label. 
The scoring function for extended typed model, \typeex, with an underlying embedding model \textbf{X} is 
\begin{align}
\begin{split}
f(s,r,o) = {}& \sigma \left(\left(\mathbf{s_t} \Vert \mathbf{s_l}\right) \cdot \left(\mathbf{r_h} \Vert \mathbf{r_{dom}}\right)\right) * \\  {}& \mathbf{Y} \left(s, r, o\right) * \\ {}&  \sigma \left( \left(\mathbf{o_t} \Vert \mathbf{o_l}\right) \cdot \left(\mathbf{r_t} \Vert \mathbf{r_{range}}\right) \right), 
\end{split}
\end{align}
where $\mathbf{s_l}$ denotes the explicit type label assigned to entity $s$, $\mathbf{r_{dom}}$ and $\mathbf{r_{range}}$ provide the explicit type labels for domain and range of a relation respectively. The type compatibility is enforced by \emph{concatenating}, denoted $\Vert$, the two vectors and taking their dot product. In case an explicit type label for an entity is unknown, we use the \emph{UNK} embedding as per the convention.
%
%

%


\section{Preparing Datasets for Evaluating the KG Refinement Task}
\label{datasets}
Before we present the details of the datasets used in our study, we first present the methodology followed to prepare them for use in the KG refinement task. 
As discussed earlier, 
apart from \nell, none of the KG benchmarks contain noise labels, making them unsuitable for evaluating the KG refinement task. We prepare them as follows:
\begin{itemize}
\item We sample a random $25\%$ of all facts (including the \texttt{typeOf} relations) and corrupt them by randomly changing their subject, relation label or object. Note that this was the same model followed in an earlier study~\cite{pujara2017sparsity}.

\item We further refine the noise model by ensuring that \emph{half} of the corrupted facts have entities that are type compatible to the relation of the fact. This makes it harder for detecting corrupted facts simply by using type compatibility checks. 
\end{itemize}

To capture realistic KG refinement settings, we further add extraction scores generated by sampling them from two different normal distributions: $N(0.7,0.2)$ for facts in the original KG and $N(0.3,0.2)$ for added noisy facts \cite{pujara2013knowledge}. The SAMEENT facts between entities are generated by calculating the average of the two Jaccard similarity score over sets of relationships with these pair of entities as head and tail entity respectively -- the average score acts as the confidence score of the fact. Finally, for all datasets, the test and validation sets are created by randomly partitioning the KG. Note that for all datasets 
the test set also includes the facts that were part of the original benchmark test collection. 

\subsection{Datasets}
\begin{description}
\item[\nell:]{The \nell subset taken from its 165$^{th}$ iteration \cite{carlson2010toward}) has been used for the KG refinement task~\cite{pujara2013knowledge,jiang2012learning}. It comes with a rich ontology from the NELL system, and contains multiple sources of information i.e., a single fact is present with multiple extraction scores.{ Since the original dataset does not have validation set, we split the test set into 2 equal halves preserving the same class balance, and use them as our validation and test split.} }

\item[\yago:]{\yago~\cite{dettmers2018convolutional} is a subset of the YAGO3 \cite{suchanek2007yago} knowledge graph. It is often used for evaluating the KG completion task. We have augmented it with ontological facts and entity types derived from YAGO3.  Since YAGO3 has a large number of types, we contract the type hierarchy to make it comparable to other datasets. We linked YAGO facts directly with the YAGO taxonomy by skipping the rdf:type entities at leaves of taxonomy (from YAGO simple types) and the first level of YAGO taxonomy. Then all facts upto length 3 in the hierarchy of taxonomy were included. } 

\item [\fb:]{\fb 
~\cite{dettmers2018convolutional}, another popular benchmark does not have ontological and type label information. Therefore, we use the type labels for entities from \cite{xie2016representation} which also provides the domain and range information for relations. The subclass information is populated by reconstructing the type hierarchy from type label facts. Mutually exclusive labels, relations and inverse relations are automatically created by mining the KG -- e.g. we can find inverse relations by checking if all reverse edges exists in the KG for a relation. }
\item [\wnet:]{\wnet, similar to \fb, does not contain ontological and type information. We used the synset information obtained from \cite{villmow18}, to assign type labels for entities. For example, for synset \textsf{hello.n.01}, the type is considered as \texttt{noun(n)}. Using an older ontology\footnote{https://www.w3.org/2006/03/wn/wn20/} we derived the rest of ontological information for the dataset. }
\end{description}

\begin{table*}[t!]
\small
\centering
\begin{tabular}{lccc}
\toprule
\textbf{Dataset} & \textbf{$|E|$} & \textbf{$|R|$} & \textbf{\#triples in train / valid / test}\\
\midrule
NELL & 820K & 222 & 1.02M / 4K / 4K  \\
FB15K-237 & 14K & 238 & 246K / 27K / 30K  \\
YAGO3-10 & 123K & 38 & 1.13M / 10K / 10K  \\
WN18RR & 40K & 12 & 116K / 6K / 6K  \\
\bottomrule
\end{tabular}
\caption{Number of entities, relation types and observed triples in datasets}
\label{kg_stats}
\end{table*}

\normalsize

\begin{table*}[t!]
\centering
\small
{\begin{tabular}{lcccccccc}
\toprule
\textbf{Dataset} & \textbf{DOM} & \textbf{RNG} & \textbf{SUB} & \textbf{RSUB} & \textbf{MUT} & \textbf{RMUT} & \textbf{INV} & \textbf{SAMEENT}\\\midrule
NELL & 418 & 418 & 288 & 461 & 17K & 48K & 418 & 8K\\
FB15K-237 & 237 & 237 & 44K & 0 & 147K & 53K & 44 & 20K\\
YAGO3-10 & 37 & 37 & 828 & 2 & 30 & 870 & 8 & 20K\\
WN18RR & 11 & 11 & 13 & 0 & 0 & 66 & 0 & 20K\\
\bottomrule
\end{tabular}}
\caption{\# of instances for each ontological component required by \pslkgi}
\label{onto_stats}
\end{table*}

\normalsize
Table \ref{kg_stats} summarizes the size of different KG datasets we use in our evaluation. Table~\ref{onto_stats} shows the amount of ontological information for each dataset. \nell and \fb  have reasonably rich ontological information compared to \yago and \wnet.

\section{Experimental Evaluation}
\label{evaluation_section}
We evaluate the performance of \typeex models in the KG refinement task, and compare them with \complex~\cite{trouillon2016complex} and \conve~\cite{dettmers2018convolutional}, two state-of-the-art KG embeddings methods, and \pslkgi. We also use \complex and \conve as base embedding models for our \typeex method to get \typeecomplex and \typeeconve respectively. 
We use a single hyper-parameter threshold as the cutoff for classifying a test triple based on the prediction score~\cite{pujara2013knowledge}. Our experiments were run on Intel(R) Xeon(R) x86-64 machine with 64 CPUs using 1 NVIDIA GTX 1080 Ti GPU. We observe the average running time with \typeecomplex to be between 25--100 minutes and with \typeeconve to be between 120--420 minutes per iteration. The increased time observed for \typeeconve experiments is because of the fact that \conve takes longer time to train than \complex\footnote{Additional scalability experiments are reported in Appendix \ref{sec:scalability}}. The hyper-parameter is tuned on the validation set and used unchanged for the test set.  We use $\phi_1=0.5$ and $\phi_2=0.75$ in Equation \ref{eqn:thresholds} as these hyperparameters were found to work across a variety of datasets.

The structure of experimental analysis we conducted are as follows:
\begin{itemize}
    \item In Section \ref{sec:main-results}, we report on the quality of embeddings generated by our \typeex methods compared with \complex, \conve and \pslkgi. In addition, we also compare our \emph{explicitly supervised} \typeex methods with the \emph{implicitly supervised} embeddings proposed by \cite{jain2018type}.
    
    \item In Section \ref{sec:feedback}, we analyse how our accuracy changes as we increase the number of feedback iterations.
    
    \item In Section \ref{sec:hyper-parameters}, we discuss the effect of the threshold parameters $t_2$ and $t_3$ on the size of the KG and the prediction accuracy.
    
    \item In Section \ref{sec:ablation-study}, we present an ablation study and analyse the impact of the various ontological rules on the accuracy.
    \comment{
    \item In Appendix \ref{sec:appendix}, we make a few additional, intriguing observations on scalability and accuracy that require further study.}
    
\end{itemize}

\noindent
\textbf{Evaluation Metric:} Our main evaluation metric is the \emph{weighted F1} (wF1)  measure. The reason for this is that in the KG refinement task, there is an imbalance in the two classes -- noisy facts and correct facts\footnote{Noisy facts are much lower in number compared to correct facts.}. Weighted F1 is defined as the individual class F1 score weighted by the number of instances per class in the test set.

\begin{equation}
        wF1 = w_1 * F1 (l_1) +  w_0 * F1 (l_0)
\end{equation}
where $w_k$ is the fraction of samples with label $k$ ($k \in \{0,1\}$ in our setting), $F1 (l_k)$ is the $F1$ score computed only for class $k$.

\subsection{Baselines}
In addition to the baselines \complex, \conve and \pslkgi, we compare our method with two other ensemble methods, described below.

\begin{description}
\item[\conve + \complex:] In the first stage, instead of using \pslkgi for predictions, we use \conve. These predictions (along with the original KG) are used as input to the second stage which used \complex. Note that this baseline combines to \emph{similar} methods.

\item[$\alpha$ -model:] This baseline is a simple score combination of two different methods (in contrast to the two stages with iterations of our method). We use the setting introduced in R-GCN (\citet{R-GCN}) to combine scores of KG embeddings and PSL-KGI methods using the equation given below:\\
\begin{equation}
        f(h,r,t)_{\alpha-model} = \alpha * f(h,r,t)_{\pslkgi} +  (1 - \alpha)*f(h,r,t)_{model}
\end{equation}
Here the hyperparameter $\alpha$ is chosen based on the validation set. The optimal $alpha$ value obtained are reported in table \ref{alpha_values} and $model$ could be either \complex or \conve.
\end{description}

\begin{table*}[t!]
\centering
\small
{\begin{tabular}{lccaccaccacca}
\toprule
\textbf{Method} & \textbf{\nell} & \textbf{\yago} & \textbf{\fb} & \textbf{\wnet}\\
\midrule
$\alpha$ - \complex & 1.0 & 0.4 & 0.7 & 0.3 \\
$\alpha$ - \conve & 1.0 & 0.4 & 0.6 & 0.9 \\
\bottomrule
\end{tabular}}
\caption{Optimal $\alpha$ values obtained based on performance on validation set}
\label{alpha_values}
\end{table*}
\normalsize

\subsection{Accuracy of \typeex}
\label{sec:main-results}
\begin{table*}[t!]
\centering
\small
\resizebox{\textwidth}{!}
{\begin{tabular}{lccaccaccacca}
\toprule
\textbf{Method} & \multicolumn{3}{c}{\textbf{\nell}} & \multicolumn{3}{c}{\textbf{\yago}} & \multicolumn{3}{c}{\textbf{\fb}} & \multicolumn{3}{c}{\textbf{\wnet}}\\
\cmidrule(lr){2-4} \cmidrule(lr){5-7} \cmidrule(lr){8-10} \cmidrule(lr){11-13}
 & +ve F1 & -ve F1 & wF1 & +ve F1 & -ve F1 & wF1 & +ve F1 & -ve F1 & wF1 &  +ve F1 & -ve F1 & wF1 \\ 
\midrule
\complex & 0.82 & 0.58 & 0.73 & 0.94 & 0.43 & 0.88 & 0.96 & 0.4 & 0.92 & 0.93 & 0.26 & 0.86 \\
\conve & 0.74 & 0.55 & 0.67 & 0.94 & 0.37 & 0.87 & 0.95 & 0.37 & 0.90 & 0.93	& 0.07 & 0.84 \\
    \pslkgi & 0.85 & \textbf{0.68} & \textbf{0.79} & 0.91 & 0.39 & 0.85 & 0.92 & 0.39 & 0.88 & 0.91 & \textbf{0.37} & 0.85 \\\midrule 
{\conve+\complex} & 0.82 & 0.58 & 0.73 & 0.95 & 0.43 & 0.89 & 0.96 & 0.39 & 0.92 & 0.93 & 0.15 & 0.85 \\ 
$\alpha$ - {\complex} & 0.85 & 0.68 & 0.79 & 0.94 & 0.50 & 0.89 & 0.96 & 0.58 & 0.93 & 0.94 & 0.24 & 0.87 \\
$\alpha$ - {\conve} & 0.85 & 0.68 & 0.79 & 0.94 & 0.41 & 0.88 & 0.95 & 0.47 & 0.92 & 0.92 & 0.34 & 0.85 \\\midrule
\textbf{\typeecomplex} & \textbf{0.86} & \textbf{0.68} & \textbf{0.79} & \textbf{0.95}	& \textbf{0.56} & \textbf{0.91} & \textbf{0.98} & \textbf{0.82} & \textbf{0.97} & 0.93 & 0.24 & 0.85 \\
\textbf{\typeeconve} & \textbf{0.86} & 0.67 & \textbf{0.79} & \textbf{0.95} & 0.47 & 0.89 & \textbf{0.98} & 0.77 & 0.96 & \textbf{0.94} & 0.31 & \textbf{0.87} \\
\bottomrule
\end{tabular}}
\caption{Overall performance of all models in KG refinement task using the best wF1 measure obtained in first 6 iterations. +ve F1 indicate the F1 score for correct facts and -ve F1 indicate F1 score for noisy facts.}
\label{f1_scores}
\end{table*}
\normalsize

Our main results are shown in Table \ref{f1_scores}. We include separate F1 measures for the two classes as well as the weighted F1 measure. This helps us analyse how well each method performs in identifying the correct (+ve) and noisy facts (-ve). From the table, we observe that our proposed combined methods \typeex consistently outperform the KG embeddings methods as well as the baseline \pslkgi. Note that \pslkgi is a formidable baseline over \nell since it contains a rich ontology.

For the positive class (correct facts), our method performs slightly better than the second best competitor, while for the negative class (noisy facts), both our methods show substantial improvements for \yago and \fb datasets, while performing on par with \pslkgi for \nell. The only dataset on which our methods fail to beat the PSL-KGI baseline is \wnet and this is because of its very limited ontology (please refer to Table \ref{onto_stats}). Further, for all datasets our \typeex methods have the best $wF1$ numbers. We have therefore validated our initial hypothesis that ontological information of high quality is tremendously helpful in improving the quality of embeddings. 

\paragraph*{Comparison with Baseline Ensemble Models.}
From Table \ref{f1_scores}, we see that \typeex models perform much better than \conve+\complex. We hypothesize that this is because \pslkgi and embeddings methods are complementary in nature. That is, \pslkgi is better at removing noisy facts, while embeddings methods are better at inferring new facts. In contrast, when we combine \complex and \conve, the resultant model cannot incorporate rich ontological information and, hence, cannot effectively remove noise from the KG. This intuition is confirmed by looking at low -ve F1 of these methods when compared to \typeex models in Table \ref{f1_scores}\footnote{For more observations regarding noise removal, please refer to Appendix \ref{sec:noise-removal}}.

We also observe that $\alpha$-models perform better than the corresponding individual methods, but not better than our \typeex methods. This observation shows that our methodology of combining the two approaches in a pipeline fashion is more powerful than a simple weighted combination of these methods. The reason is that, in our method, each of the individual methods benefits from the strength of the other method since the results of one are used as input for the other. As a result, both these methods gain from each other's performance. Further, we list the computed alpha values that showed the best validation performance in Table \ref{alpha_values}. We observe that the alpha values are mostly tilted towards the better-performing model.

\paragraph*{Comparison with unsupervised type inference.} 
In Table \ref{typed_model_comp} we compare the performance of \typeecomplex which has \emph{explicit} type supervision with the \emph{unsupervised} type-compatible embeddings-based method proposed by Jain et al.~\cite{jain2018type}. As these results indicate, while explicitly ensuring type compatibility helps to improve performance, adding type inferences from \pslkgi to \typeecomplex significantly improves the relation scores, improving weighted F1 up to $18\%$ (over NELL).

\begin{table}[h]
\centering
\small
{\begin{tabular}{lca}
\toprule
 Dataset & \cite{jain2018type} & \typeecomplex \\
\midrule
NELL & 0.60 & {\bf 0.71} \\
YAGO3-10 & 0.88& {\bf 0.92} \\
FB15K-237 & 0.93 & {\bf 0.97} \\
WN18RR & 0.85 & {\bf 0.85} \\
\bottomrule
\end{tabular}}
\caption{Weighted F1 scores on relation triples in the test set by \cite{jain2018type} and \typeecomplex.}
\label{typed_model_comp}
\end{table}

\normalsize
\paragraph{Anecdotes.} Looking at the example predictions by both \typeecomplex and ComplEx on YAGO3-10, we observed that \typeecomplex is able to correctly identify simple noisy facts like  {$\langle$\texttt{Leinster\_Rugby, hasGender, Republic\_of\_Ireland}$\rangle$ }, where there is a clear type incompatibility, which \complex is unable to identify. Further \typeecomplex is able to identify noisy facts such as {$\langle$\texttt{Richard\_Appleby, playsFor, Sporting\_Kansas\_City}$\rangle$}, with type compatible entities, by finding the reasoning context that connect $\texttt{Richard\_Appleby}$ with football teams of UK and not US.

\subsection{Analysis of feedback iterations}
\label{sec:feedback}

We have already shown in Table \ref{f1_scores} that our \typeex methods output higher quality predictions compared to the other baselines. In this section, we analyse the conditions under which multiple iterations can improve the quality of predictions.

\begin{figure}
    \centering
    \begin{minipage}{0.49\textwidth}
        \centering
        \includegraphics[width=\textwidth]{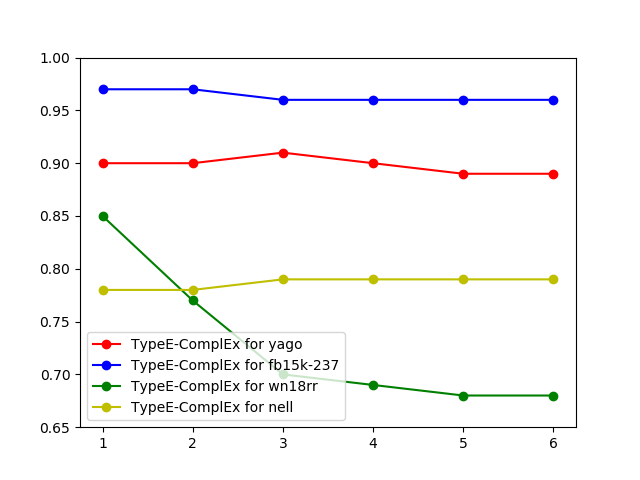}
    \end{minipage}%
    \begin{minipage}{0.49\textwidth}
        \centering
        \includegraphics[width=\textwidth]{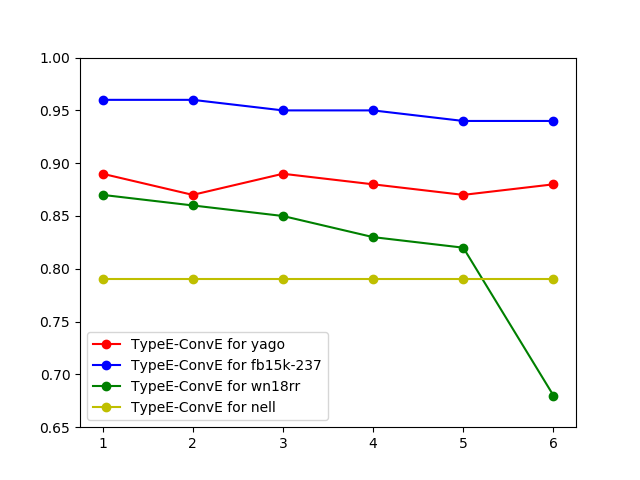}
    \end{minipage}
    \caption{Graph showing the weighted F1 score (y-axis) obtained at the given number of feedback iterations (x-axis).}
    \label{fig:feedback-iterations}
\end{figure}

Figure \ref{fig:feedback-iterations} shows how the $wF1$ values of our \typeex methods change over six feedback iterations. Recall from Figure \ref{combined_model} that each iteration involves adding high quality tuples from \pslkgi inferences to \typeex and feeding back high quality tuples from \typeex predictions back to \pslkgi.

The main observation we make in Figure \ref{fig:feedback-iterations} is that the accuracy of predictions on datasets with a rich and good quality ontology (\nell, \fb and \yago) do not do not vary much. In fact, for \nell and \yago the accuracy actually increases in multiple iterations (best accuracy for \nell is in the $6^{th}$ iteration, and for \yago it is in the $3^{rd}$), while for \fb, there is only a small decrease over the first and last iterations.

In contrast, for the \wnet dataset, the accuracy degrades quite rapidly after the first iteration. The reason is that this dataset does not have even a moderate number of ontological rules that are of high quality\footnote{Recall from Section \ref{datasets} that the ontology rules were obtained from \cite{villmow18} and an older ontology.}. This results in lower quality inference from \pslkgi which feeds into the \typeex method. This results in lower quality predictions from \typeex, which is then fed back into \pslkgi. Thus a cascading effect of low quality predictions from each method results in a rapid drop in prediction quality.

\subsection{Impact of hyper-parameters $t_1$ and $t_2$}
\label{sec:hyper-parameters}

\begin{figure}[ht]
    \centering
    \begin{minipage}{0.4\textwidth}
        \centering
        \includegraphics[width=\textwidth]{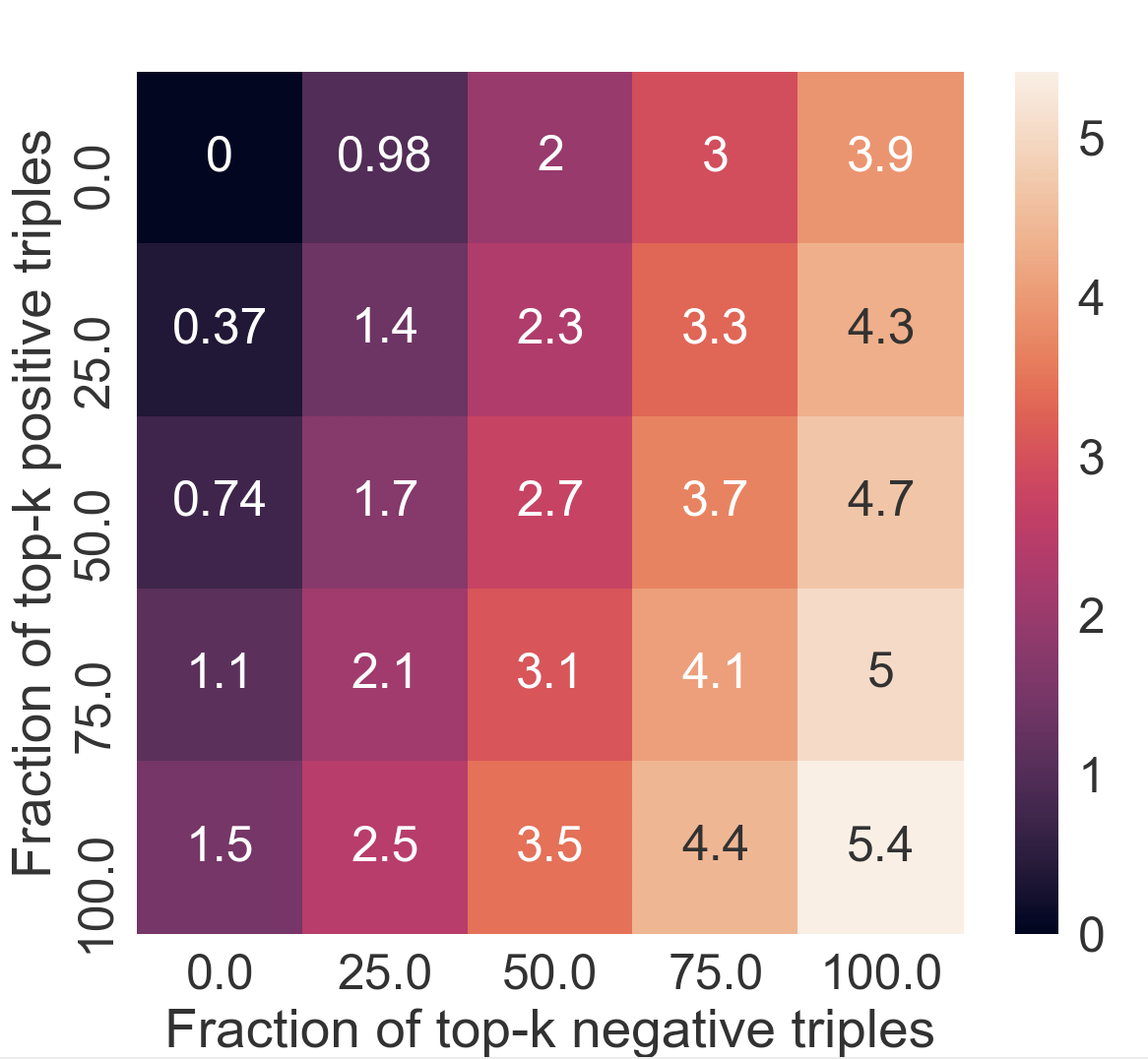}
    \end{minipage}%
     \begin{minipage}{0.4\textwidth}
        \centering
        \includegraphics[width=\textwidth]{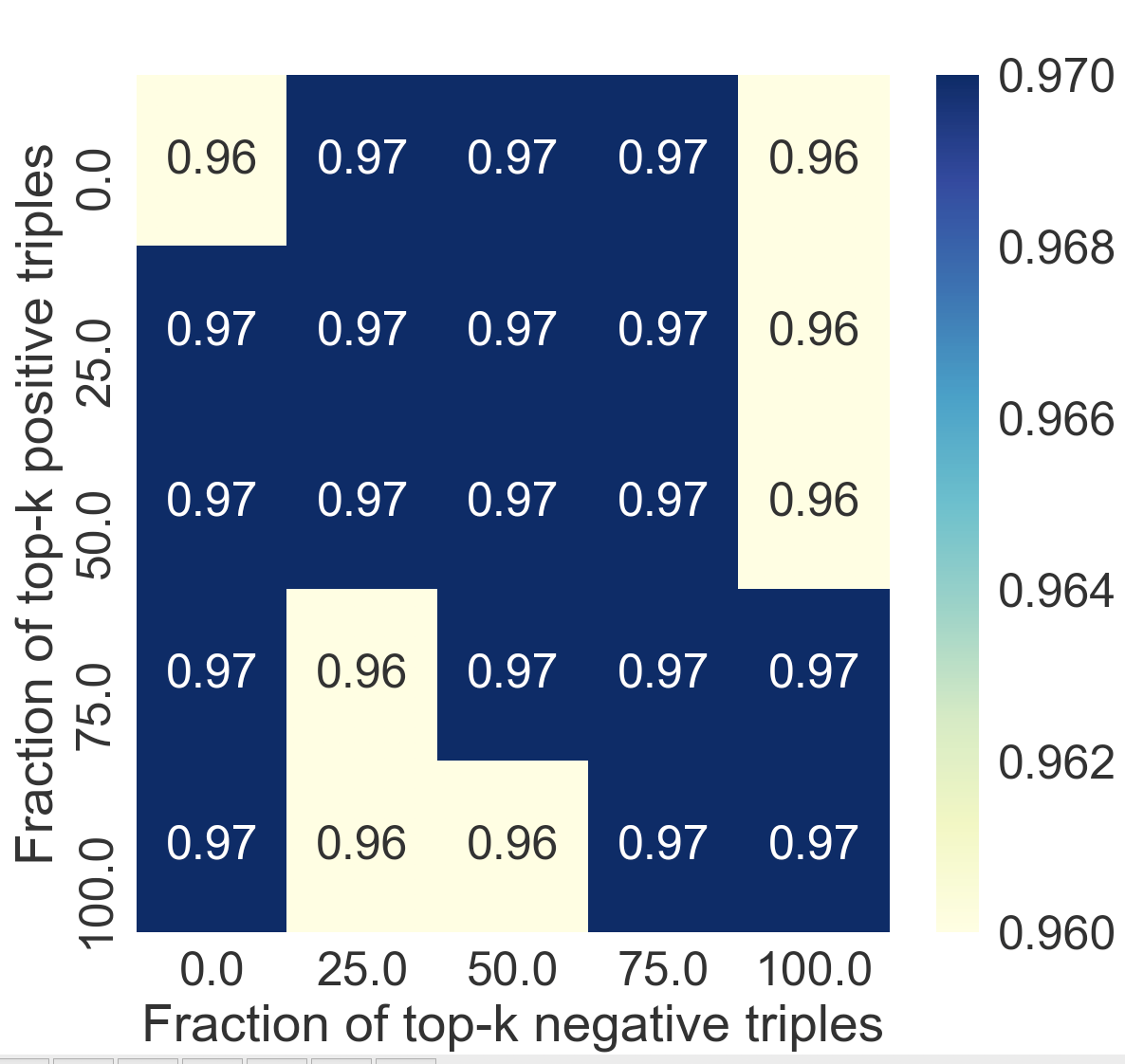}
    \end{minipage}
    \caption{(Left) Variation of size with percentage of top positive and negative triples for \typeecomplex after the first feedback iteration. (Right) Variation of wF1. Both heatmaps are for \fb}
    \label{fig:gridgraphs}
\end{figure}

 The threshold parameters $t_1$ and $t_2$ determine how many positive and negative triples are fed back to \pslkgi from \typeex. The number of such feedback triples has an impact on both, the size of the KG as well as the accuracy of predictions (because \pslkgi now performs inference using the new triples that have been fed back). Figure \ref{fig:gridgraphs} shows, for \fb, two heatmaps which quantify the impact of $t_1$ and $t_2$. In the left heatmap, the impact of adding the top-$k$ percent of positive and negative tuples on the \emph{size} of the KG is shown\footnote{The size is normalized: $\frac{(new size - original size)}{(original size)}$} and in the right heatmap, the impact on the \emph{accuracy} is shown. We observe that by adding very few positive and negative tuples, with slightly more positive tuples than negative tuples as feedback is sufficient to obtain the best accuracy, while ensuring that the KG size does not explode. 

\subsection{Ablation Study}
\label{sec:ablation-study}

\comment{
\begin{table*}[t!]
\centering
{\begin{tabular}{l*{8}{c}}
\toprule
\textbf{Method} & \multicolumn{4}{c}{\textbf{NELL}} & \multicolumn{4}{c}{\textbf{Fb15k-237}} \\
\cmidrule(lr){2-5} \cmidrule(lr){6-9}
 & wf1 & size & -ve F1 & +ve F1 & wf1 & size & -ve F1 & +ve F1 \\ 
\midrule
All rules & 0.79 & 0.48	& 0.68 & 0.86 &	0.97 & 1.74 & 0.8 & 0.98 \\
w/o Domain & 0.78 & 0.63 & 0.65 & 0.85 & 0.96 & 1.62 & 0.76 & 0.98\\
w/o Same Entity & 0.79	& 0.17 & 0.67 & 0.85 & 0.97 & 1.02 & 0.8 & 0.98\\
w/o MUT & 0.79 & 0.34 & 0.68 & 0.85 & 0.97 & 1.37 & 0.8 & 0.98 \\
w/o Range & 0.76 & 0.42 & 0.65 & 0.82 & 0.95 & 1.89 & 0.72 & 0.97 \\
w/o Subclass & 0.77 & 0.55 & 0.63 & 0.84 & 0.97 & 2.56 & 0.81 & 0.98\\
w/o RMUT & 0.79 & 0.49 & 0.67 & 0.86 & 0.97 & 1.51 & 0.8 & 0.98\\
w/o INV & 0.78 & 0.3 & 0.66 & 0.85 & 0.97 & 1.09 & 0.81 & 0.98\\
w/o SUBPROP & 0.79 & 0.07 & 0.67 & 0.86 & 0.97 & 1.74 & 0.8 & 0.98\\
\bottomrule
\end{tabular}}
\caption{Ablation study for performance without ontology subclass in KG refinement task for \typeecomplex models. For brevity, we have shown results for fb15k-237 at end of second epoch and NELL at end of third epoch. Size is normalised by number of triples in original KG.\textcolor{blue}{The difference of the ablated results, to the overall (All) are also mentioned in bracket along with the actual numbers.}}
\label{ontology_table}
\end{table*}
}

\begin{table*}[t!]
\centering
{\begin{tabular}{l*{6}{c}}
\toprule
\textbf{Method} & \multicolumn{3}{c}{\textbf{\nell}} & \multicolumn{3}{c}{\textbf{\fb}} \\
\cmidrule(lr){2-4} \cmidrule(lr){5-7}
 & +ve F1  & -ve F1 & wf1 & +ve F1 & -ve F1 &  wF1 \\ 
\midrule
All rules       & \textbf{0.86} & \textbf{0.68} & \textbf{0.79} & \textbf{0.98} & 0.80 & \textbf{0.97} \\
No rules       & 0.82 & 0.58 & 0.73 & 0.96 & 0.4 & 0.92 \\
\cmidrule{1-7}
w/o DOM      & 0.85 (-0.01) & 0.65 (-0.03) & 0.78 (-0.01) & \textbf{0.98} (0.00) & 0.76 (-0.04) & 0.96 (-0.01) \\
w/o SAMEENT & 0.85 (-0.01) & 0.67 (-0.01) & \textbf{0.79} (0.00) & \textbf{0.98} (0.00) & 0.80 (0.00) & \textbf{0.97} (0.00) \\
w/o MUT & 0.85 (-0.01) & \textbf{0.68} (0.00) & \textbf{0.79} (0.00) & \textbf{0.98} (0.00) & 0.80 (0.00) & \textbf{0.97} (0.00) \\
w/o RNG   & 0.82 (-0.04) & 0.65 (-0.03) & 0.76 (-0.03) & 0.97 (-0.01) & 0.72 (-0.08) & 0.95 (-0.02) \\
w/o SUB    & 0.84 (-0.02) & 0.63 (-0.05) & 0.77 (-0.02) & \textbf{0.98} (0.00) & \textbf{0.81} (0.01) & \textbf{0.97} (0.00) \\
w/o RMUT    & \textbf{0.86} (0.00) & 0.67 (-0.01) & \textbf{0.79} (0.00) & \textbf{0.98} (0.00) & 0.80 (0.00) & \textbf{0.97} (0.00) \\
w/o INV       & 0.85 (-0.01) & 0.66 (-0.02) & 0.78 (-0.01) & \textbf{0.98} (0.00) & \textbf{0.81} (0.01) & \textbf{0.97} (0.00) \\
w/o RSUB     & \textbf{0.86} (0.00) & 0.67 (-0.01) & \textbf{0.79} (0.00) & \textbf{0.98} (0.00) & 0.80 (0.00) & \textbf{0.97} (0.00) \\
\cmidrule{1-7}
ONLY DOM+RNG      & 0.84 (-0.02) & 0.65 (-0.03) & 0.77 (-0.02) & \textbf{0.98} (0.00) & 0.80 (0.00) & \textbf{0.97} (0.00) \\
ONLY DOM      & 0.84 (-0.02) & 0.64 (-0.04) & 0.77 (-0.02) & \textbf{0.98} (0.00) & 0.73 (-0.07) & 0.96 (-0.01) \\
ONLY RNG      & 0.83 (-0.03) & 0.63 (-0.05) & 0.76 (-0.03) & \textbf{0.98} (0.00) & 0.76 (-0.04) & 0.96 (-0.01) \\
ONLY SAMEENT & 0.83 (-0.03) & 0.63 (-0.05) & 0.76 (-0.03) & \textbf{0.98} (0.00) & 0.73 (-0.07) & 0.96 (-0.01) \\
ONLY MUT & 0.83 (-0.03) & 0.63 (-0.05) & 0.76 (-0.03) & \textbf{0.98} (0.00) & 0.73 (-0.07) & 0.96 (-0.01) \\
ONLY SUB    & 0.82 (-0.04) & 0.60 (-0.08) & 0.74 (-0.05) & \textbf{0.98} (0.00) & 0.74 (-0.06) & 0.96 (-0.01) \\
ONLY RMUT  & 0.83 (-0.03) & 0.62 (-0.06) & 0.76 (-0.03) & \textbf{0.98} (0.00) & 0.76 (-0.04) & 0.96 (-0.01) \\
ONLY INV & 0.84 (-0.02) & 0.63 (-0.05) & 0.76 (-0.03) & \textbf{0.98} (0.00) & 0.73 (-0.07) & 0.96 (-0.01) \\
ONLY RSUB     & 0.83 (-0.03) & 0.62 (-0.06) & 0.76 (-0.03) & - & - & - \\
\bottomrule
\end{tabular}}
\caption{Ablation study for performance without ontology subclass in KG refinement task for \typeecomplex models. We have shown results for \fb at end of second epoch and \nell at end of third epoch.}
\label{ontology_table}
\end{table*}

We performed an ablation study to determine what kind of ontology rules were most useful in increasing prediction accuracy. The results for two datasets, \nell and \fb are shown in Figure \ref{ontology_table}. From the table, we observe that the it is the Subclass, Domain and Range rules that are the most important. Clearly these rules are most useful in correctly predicting types, which in turn are crucial for the accuracy of the \typeex methods.

Further, as these results show, none of the individual ontological components alone show performance comparable to using all the components (and thus all the rules) in the PSL-KGI phase of \ourm. Although positive class performance over \fb remains unchanged when using \emph{any} one ontological component, the performance over negative classes deteriorates significantly over using all the components. Thus, we argue that our proposal of using as much ontological information available in a KG is consistently superior for the KG refinement task.



\section{Conclusion and Future work}
We have looked at the KG refinement task and methods for the same, from probabilistic rule based methods like PSL-KGI \cite{pujara2013knowledge} to KG embedding methods like type-ComplEx \cite{jain2018type}. We showed their performance on existing datasets in the literature, and how the extent pf ontology plays a crucial role in performance of PSL-KGI. 

To overcome their individual limitations, we present a simple mechanism to combine the PSL-KGI and typed models, by providing feedback inputs to each of the two methods. Such a method provides the individual model components with additional context, generated from each other, which leads to an increase in performance of both components on almost all datasets used for the KG refinement task evaluation. We further closely inspect into what controls the magnitude of the improvement and what are effect of further iterations of this feedback mechanism.

For future work, we will look in ways to training the whole pipeline end to end thus increasing the flexibility and efficieny of our approach. Further we will also look into the concept of data augmentation for KG, so that more context is available related to a triple for prediction.

\pagebreak
\bibliographystyle{plainnat}
\bibliography{biblo}
\pagebreak
\appendix

\section{}
\label{sec:appendix}
\begin{figure}[ht]
    \centering
    \begin{minipage}{0.4\textwidth}
        \centering
        \includegraphics[width=\textwidth]{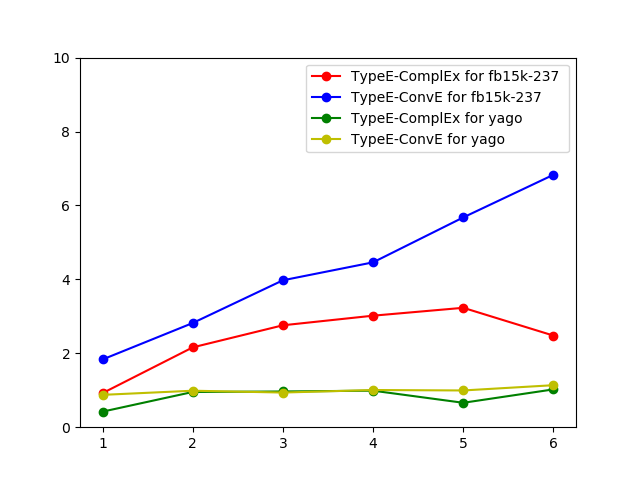}
    \end{minipage}%
    \caption{Graph showing the Knowledge Graph size (y-axis) obtained at the given number of feedback iterations (x-axis). Size is normalised by number of triples in original KG.}
    \label{fig:sizegraphs}
\end{figure}

\begin{figure}
    \centering
    \begin{minipage}{0.4\textwidth}
        \centering
        \includegraphics[width=\textwidth]{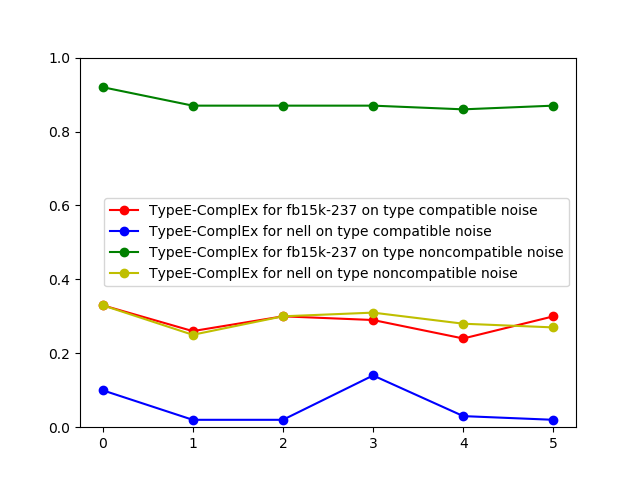}
    \end{minipage}
    \caption{Graph showing the variation of recall (y-axis) on added noise with increasing number of iterations (x-axis).}
    \label{fig:typegraphs}
\end{figure}
\subsection{Scalability issues}
\label{sec:scalability}

Scalability is an essential issue in the feedback experiments. We initially passed the predictions of \typeex embedding on the entire knowledge graph as feedback to the \pslkgi model. We observed that this approach does not make the model scalable, and the memory required by  \pslkgi increased exponentially. With our current approach using thresholds $t_2$ and $t_3$ in order to feed back only the high confidence triples, we are able to prevent the size of the KG from exploding. Figure \ref{fig:sizegraphs} shows the size of the KG for the various datasets over different iterations.
We observe two key behaviors, as seen in Figure ~\ref{fig:sizegraphs}. We see that for \fb, the size almost uniformly increases with increasing iterations, whereas, for \yago, the size of KG becomes stable after a few iterations. This phenomenon is again related to how useful the initial ontology is to help the \pslkgi model to filter noise added at every iteration. Since for Yago and NELL, ontology comes with the datasets, they are of high quality, and we observe much more stability in Knowledge Graph sizes.

\subsection{Noise removal}
\label{sec:noise-removal}
We investigate our performance on cleaning up noisy KGs. The observed behavior can be seen in figure ~\ref{fig:typegraphs}. As expected, type compatible noise is harder to remove than type noncompatible noise.  Moreover, for both datasets, the performance on type compatible noise seems to get better for the first few iterations. This can be attributed to the fact that feedback in terms of type predictions and high-quality relation triples from \pslkgi are getting better with the increasing number of iterations. This improvement shows that the feedback sent from \typeex models seems to be very helpful to \pslkgi model thus motivating our approach. Finally, our model seems to be stable in its performance on noise removal with increasing iterations.

\subsection{\ourm Algorithm}
\label{sec:psuedo_algo}
\begin{algorithm}[h!tb]

  \caption{\ourm}\label{euclid}
  \begin{algorithmic}[1]
    \Procedure{\typeex}{$KG,inst$}\LineComment{\texttt{Algorithm on input KG with ontological instances}}
      \For{\text{$iter$ in [$1,MAX\_ITER$]}}
          \LineComment{\texttt{Predictions are for existing \& inferred triplets and type labels. $Rel\_Pred$ denotes relation prediction and $Type\_pred$ denote type predictions.}}
          \State $\text{($Rel\_Pred, Type\_pred$)} \gets \text{\pslkgi($KG,inst$)}$ 
          \For{\text{$thresh$ in [0,1]}}
            \LineComment{\texttt{Use the threshold to classify triplets as noisy or valid}}
            \State $\text{($Rel\_Labels, Type\_Labels$)} \gets \text{($Rel\_Pred > thresh, Type\_pred > thresh$)}$ 
            \LineComment{\texttt{Compute Validation Performance on KG Refinement}}
            \State $\text{Valid\_Performance} \gets \text{Test($Valid\_Rel\_Pred, Valid\_Type\_pred$)}$ 
          \EndFor
          \LineComment{\texttt{Use the threshold that maximises performance on validation set }}
          \State $t_{best} \gets max_{thresh}\text{(Valid Performance)}$ 
          \LineComment{\texttt{Predictions having less than $t_{best}$ removed}}
          \State $\text{$KG\_clean$} \gets \text{FILTER($KG$,} t_{best})$ 
          \LineComment{\texttt{Predict type for each entity by using type label prediction from \pslkgi and using subclass constraints to contract type hierarachy}}
          \State $\text{$Types$} \gets \text{GENERATE\_TYPE ($Type\_pred, SUB\_inst$)}$
          \LineComment{\texttt{Use the cleaner Knowledge Graph and high quality type predictions to train KG embedding method}}
          \State $\text{$MODEL$} \gets\text{\typeex($KG\_clean, Types$)}$
          \LineComment{\texttt{Obtained in fashion similar to $t_{best}$}}
          \State $t_{1} \gets\text{Best validation performance for $MODEL$}$ 
          \LineComment{\texttt{Follow Equation \ref{eqn:thresholds} to send some of the triplets along with their predicted score as feedback for \pslkgi method}}
          \State $Infer \gets \text{INFER\_NEW\_TRIPLES ($MODEL, KG\_clean$)}$
          \LineComment{\texttt{Evaluate \typeex methods on test set}}
          \State $\text{Test Performance} \gets \text{PREDICT ($MODEL, TEST SET$)}$ 
          \LineComment{\texttt{Update the KG with the feedback from \typeex methods}}
          \State $KG \gets KG\_clean\cup Infer$ 
      \EndFor
          
    \EndProcedure
  \end{algorithmic}
\end{algorithm}


\end{document}